\documentclass[journal]{IEEEtran}

\usepackage{cite}
\usepackage{amsmath}
\usepackage{color}
\usepackage{booktabs}
\usepackage[dvipsnames]{xcolor}
\usepackage{xspace}
\usepackage{kotex}
\usepackage[pdftex]{graphicx}
\graphicspath{{./figures/}}

\ifCLASSINFOpdf
\else
\fi

\usepackage{url}

\newcommand{\fref}[1]{Fig. \ref{#1}}
\newcommand{\tref}[1]{Table \ref{#1}}

\newcommand*{\eg}{e.g.\@\xspace}
\newcommand*{\ie}{i.e.\@\xspace}

\begin{document}
\title{Deep Dense Local Feature Matching and \\ Vehicle Removal for Indoor Visual Localization}
%
%
%

\author{
        Kyung~Ho~Park*
\thanks{*Kyung Ho Park is with SOCAR AI Research.}
}

%
%

%

\maketitle

\begin{abstract}
Visual localization is an essential component of intelligent transportation systems, enabling broad applications that require understanding one’s self-location when other sensors are not available. It is mostly tackled by image retrieval such that the location of a query image is determined by its closest match in the pre-collected images. Existing approaches focus on large-scale localization where landmarks are helpful in finding the location. However, visual localization becomes challenging in small-scale environments where objects are hardly recognizable. In this paper, we propose a visual localization framework that robustly finds the match for a query among the images collected from indoor parking lots. It is a challenging problem when the vehicles in the images share similar appearances and are frequently replaced, e.g., parking lots. We propose to 1) employ a deep dense local feature matching that resembles human perception to find correspondences and 2) eliminate matches from vehicles automatically with a vehicle detector. The proposed solution is robust to the scenes with low textures and invariant to false matches caused by vehicles. We compare our framework with alternatives to validate our superiority on a benchmark dataset containing 267 pre-collected images and 99 query images taken from 34 sections of a parking lot. Our method achieves 86.9\% accuracy, outperforming the alternatives. 
\end{abstract}

\begin{IEEEkeywords}
Indoor visual localization, Textureless environment, Deep dense local feature matching, Vehicle removal
\end{IEEEkeywords}

%
\IEEEpeerreviewmaketitle

\section{Introduction}

\IEEEPARstart{V}{isual}
localization is to find the place where the query image is taken. It enables location-based services such as navigation systems and recommending nearby attractions. It is an essential component of intelligent transportation systems, especially when other sensors are not available. The common practice of visual localization considers the problem as matching the query image to the pre-collected images so that the system produces the annotated location of the closest image.

\begin{figure}[t]
    \includegraphics[width=\linewidth]{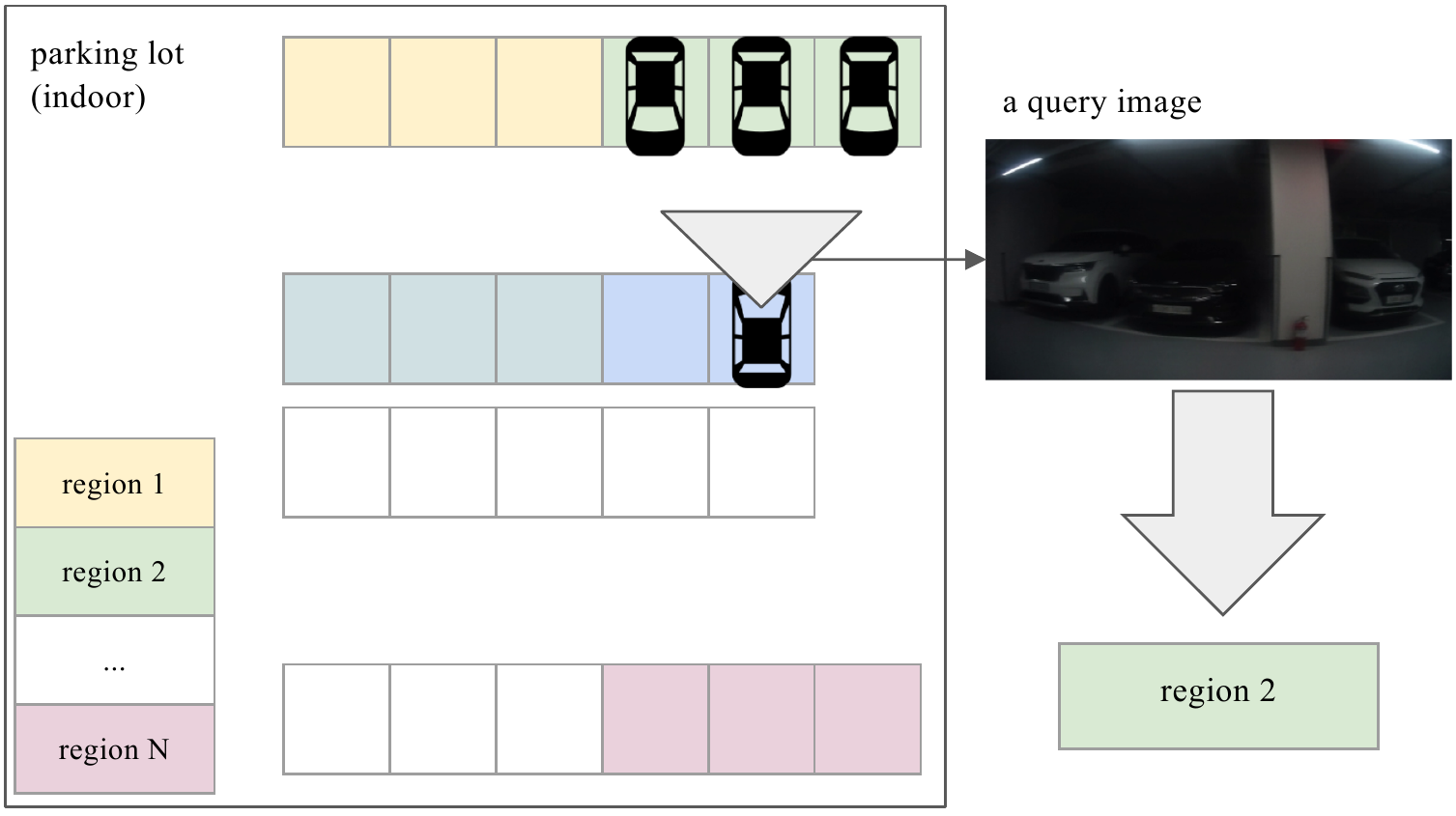}
    \caption{
    \textbf{The goal of visual localization} is to localize a query image among pre-collected images. As the locations of the individual pre-collected images are annotated, the visual localization becomes image retrieval.
    }
    \label{fig:problem}
\end{figure}

\fref{fig:problem} illustrates our problem setting. We assume that the images of an indoor parking lot are taken by a front-looking camera on a vehicle parked on a section, and the annotators have assigned the sections to the images. Given a query image from another parked vehicle, we predict the location where the vehicle is parked, marked by the section IDs. Visual localization in parking lots is a challenging problem in two senses: the walls are texture-less and vehicles act as noise. The texture-less walls make finding correspondence difficult. Multiple identical-looking vehicles are likely to appear and vehicles can move or even disappear.

\begin{figure*}[t]
    \includegraphics[width=\linewidth]{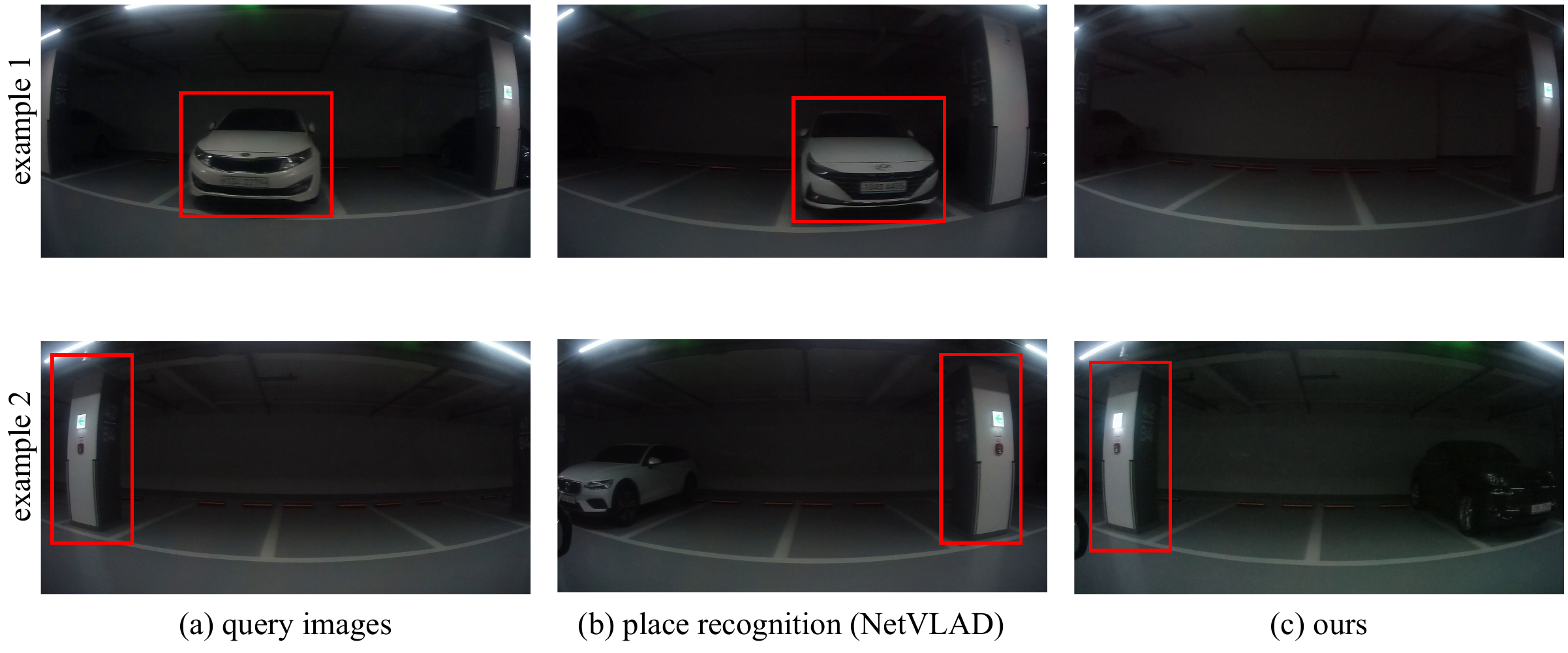}
    \caption{
    \textbf{Limitation of the NetVLAD-based approaches.} NetVLAD is prone to false matches due to the salient objects in the scenes. First row: the cars in the red boxes lead to false matches in NetVLAD. Second row: the lights in the NetVLAD result has different symbol from the query (zoom in to compare).
    }
    \label{fig:netvlad}
\end{figure*}

Early approaches regard visual localization \cite{arandjelovic2016netvlad} as recognizing landmarks in the scene because the global coordinates of the landmarks are known and the landmarks are unique. However, such approaches are not applicable in our problem because the two conditions are not met: objects repeatedly occur and are replaceable (\fref{fig:netvlad}). On the other hand, matching feature points is not restricted to recognizing unique landmarks, but is able to find correspondences between ordinary images.

\begin{figure}[t]
    \includegraphics[width=\linewidth]{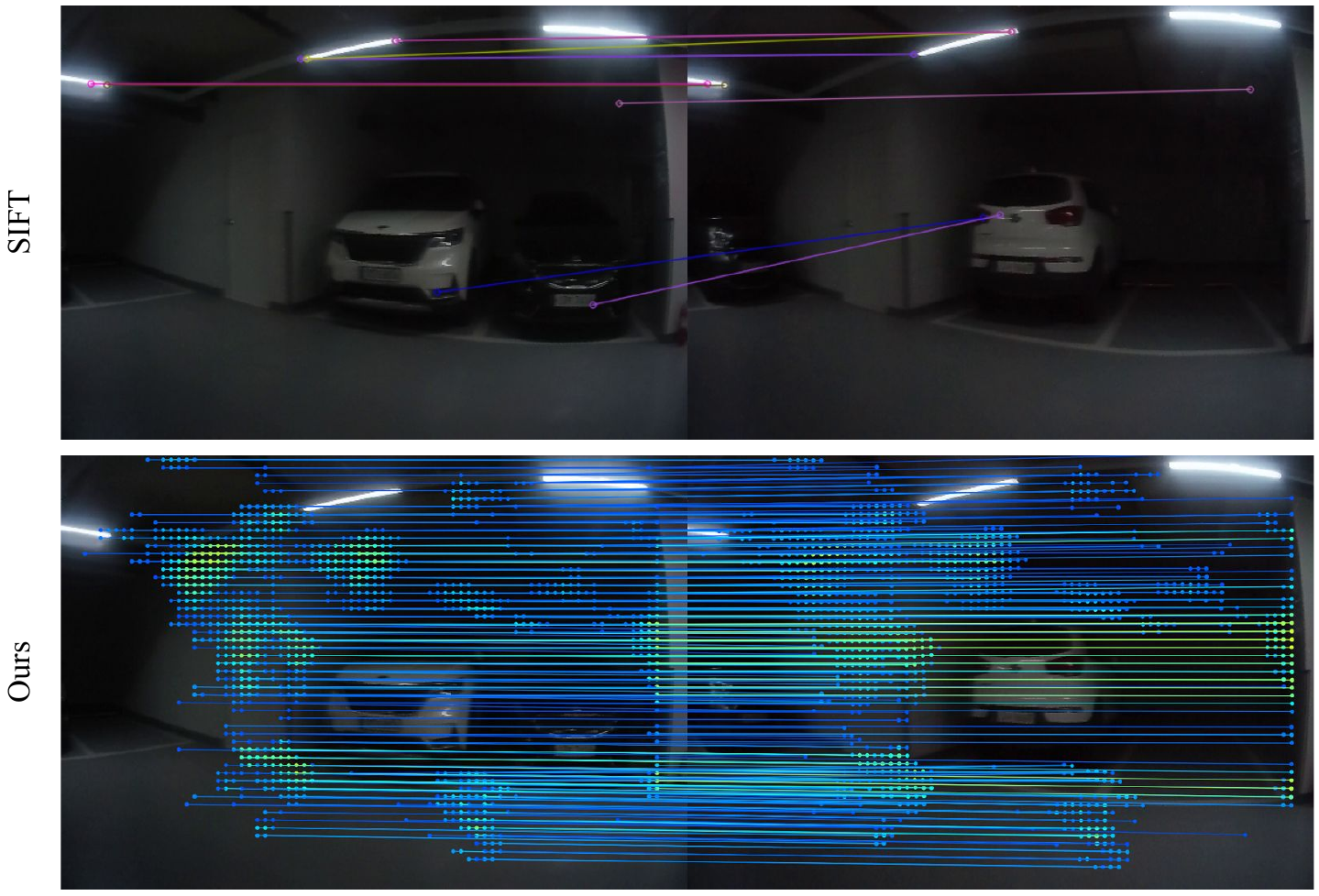}
    \caption{
    \textbf{Failure cases of the sparse matching methods.} Sparse matching methods rely on interest points that have distinctive appearances. They tend to suffer from textureless walls in indoor parking lots.
    }
    \label{fig:sparse}
\end{figure}

Finding sparse correspondences consists of three steps: 1) detecting interest points, 2) extracting descriptors, and 3) matching the points. Most existing interest point detectors \cite{lowe2004distinctive,bay2006surf, calonder2010brief, rublee2011orb, detone2018superpoint, sarlin2020superglue} struggle in our problem because the scenes are poorly textured, e.g., painted walls. \fref{fig:sparse} provides examples where sparse correspondences cannot be found. On the other hand, dense matching approaches employ correlation-based techniques which are robust to poorly-textured environments. 

To this end, we propose a framework with a deep dense local feature matching that resembles human perception to find correspondences and 2) eliminates matches from vehicles automatically with a vehicle detector. The deep dense matcher plays an important role, providing reliable correspondences even in the textureless regions. The vehicle remover handles the overwhelming number of matches on identical-looking vehicles from different locations. These two aspects pose our solution as superior to place recognition approaches and SLAM approaches in parking lot scenarios.

The test dataset contains 267 images taken from a parking lot which is split into 34 sections. The sections are manually annotated to the images by humans. The query set consists of 99 images, each of which is labeled to the corresponding section as well.
The dataset will be publicly available.

Our method is robust to the environments with low textures and to the false matches caused by repeating identical vehicles.  Comparison to the alternative choices verifies the effectiveness of our method. The code will be released for future researchers.

\section{Related work}
In this section, we briefly summarize existing methods and describe the necessity of our framework.

\subsection{Place recognition}
Approaches with vectors of locally aggregated descriptors (VLAD) \cite{jegou2011aggregating} based on deep neural networks \cite{arandjelovic2016netvlad,yu2019spatial,hausler2021patch} have achieved impressive performance in place recognition. They use a CNN pre-trained for image classification to extract feature maps whose per-pixel vectors are interpreted as local descriptors. Then, the local descriptors are divided into a fixed number of clusters and each cluster produces a vector by weighted accumulation. However, this approach may fail when the environment does not have distinctive objects such as indoor parking lots. The same issue arises in CNN-based place recognition models such as \cite{chen2014convolutional} and \cite{sunderhauf2015place}. On the other hand, our framework finds correspondences on textureless scenes and filter out correspondences on vehicles to prevent false matches.

\subsection{Point matching}

\paragraph{Handcraft features.} SIFT and SURF are commonly used traditional algorithms for extracting distinctive features in the given images. Since they are manually engineered, they offer the advantage of not requiring any training procedures or datasets. These methods are insensitive to geometric transformation such as rotation or scaling. However, they are weak to illumination changes or low textures, which is problematic in indoor parking lots. Thus, we consider recent deep learning-based approaches robust to various factors.

\paragraph{Deep features.} Recent deep neural networks have shown promising results in feature extraction. The extracted features can be used for finding correspondences between images. TimeCycle \cite{wang2019learning} and VideoWalk \cite{jabri2020walk} learn correspondence from the cycle-consistency of time. Although they achieve reasonable performance on images with distinctive local appearances, they suffer from uncertainty of cycle-consistency on homogeneous regions. On the other hand, LoFTR \cite{sun2021loftr} produces dense matches even in the low-textured areas, thanks to positional encoding. Therefore, we adopt LoFTR as a local feature matcher.

\begin{figure*}[t]
    \includegraphics[width=\linewidth]{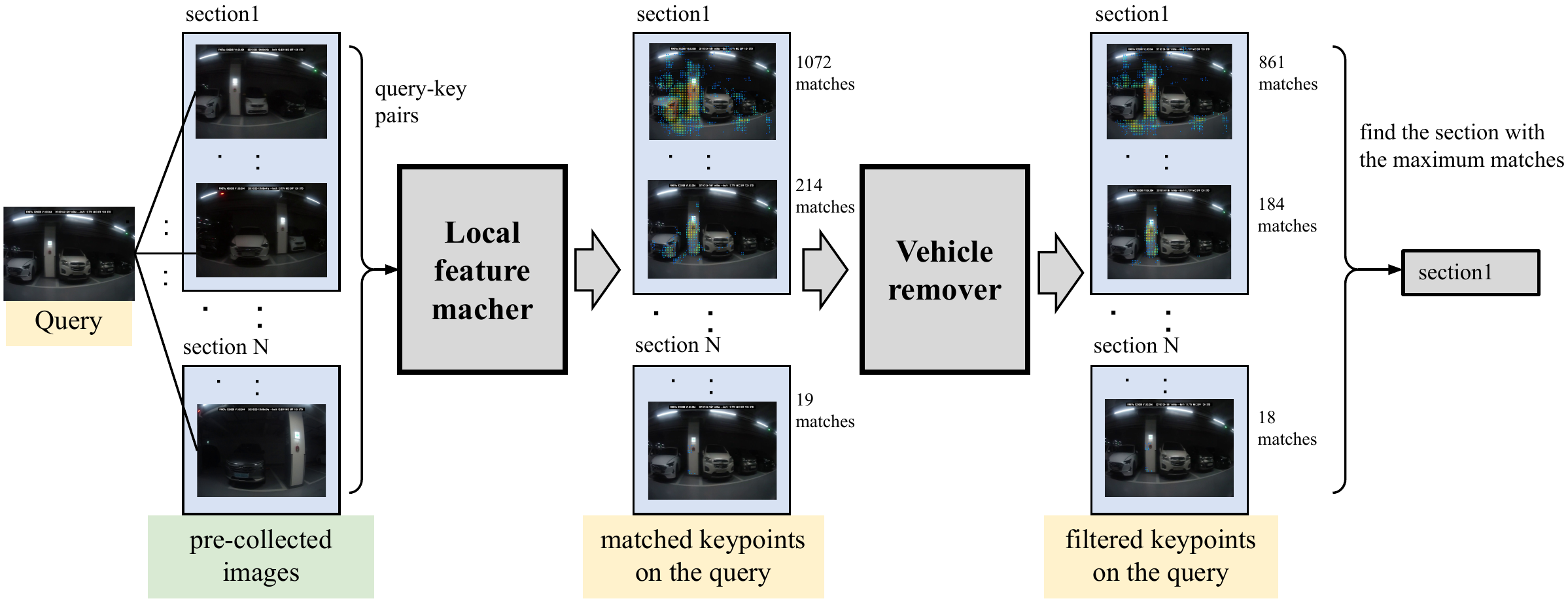}
    \caption{
    \textbf{Framework overview.} Our framework consists of two main components: a local feature matcher and a vehicle remover. The local feature matcher finds corresponding points between a query image and pre-collected images. The vehicle remover filters out confusing correspondences on vehicles. Then the section with the most correspondence becomes our prediction.
    }
    \label{fig:overview}
\end{figure*}

\subsection{SLAM and image retrieval}
Recent methods on simultaneous localization and mapping (SLAM) \cite{davison2003real, salas2013slam++, mur2015orb} have achieved impressive performance in real-time scenarios, even with dynamic environments. However, SLAM works on calibrated cameras and sequence of images, which are impractical in in-the-wild settings and the limited number of images, respectively. In this work, we tackle visual localization of parked vehicles on parking lots; the vehicles have different camera settings and the number of query images is limited to one, \ie, only the last image is available taken just before the vehicle shuts down. Hence, we exclude SLAM and rely on point-matching.

Image retrieval \cite{babenko2015aggregating, gordo2016deep} has a similar goal to visual localization: finding matches of a query image from a set of images. However, it considers diverse aspects of images such as salient objects, color distribution, styles, and semantics. On the other hand, visual localization aims to find the exactly same observations but maybe in different conditions. Hence, we also exclude image retrieval approaches.

%
%

\section{Proposed framework}
In this section, we describe our problem setting and framework to solve the problem.

\subsection{Overview}
We formulate the visual localization problem as finding the closest match of a query image among a set of stored images. It is a reasonable choice because the stored images can be paired to their locations on the map in advance, e.g., a section in a parking lot.
Our framework consists of matching, removing vehicles, and localization. The matcher finds dense correspondences between two images. The vehicle remover discards matches on vehicles. Given the filtered matches, the query image is predicted to be the same section of the image with the most matches. \fref{fig:overview} overviews our framework.

\subsection{Deep local feature matching}
Local feature matching is to find corresponding points between two images. Most traditional methods \cite{lowe2004distinctive, bay2006surf} consist of three separate phases: detecting interest points, building descriptors, and matching. Detecting interest points approaches struggle when the scenes are poorly textured. Hence, we employ deep dense local feature matching \cite{sun2021loftr} which does not require detecting interest points.

\begin{figure*}[t]
\begin{center}
    \includegraphics[width=0.75\linewidth]{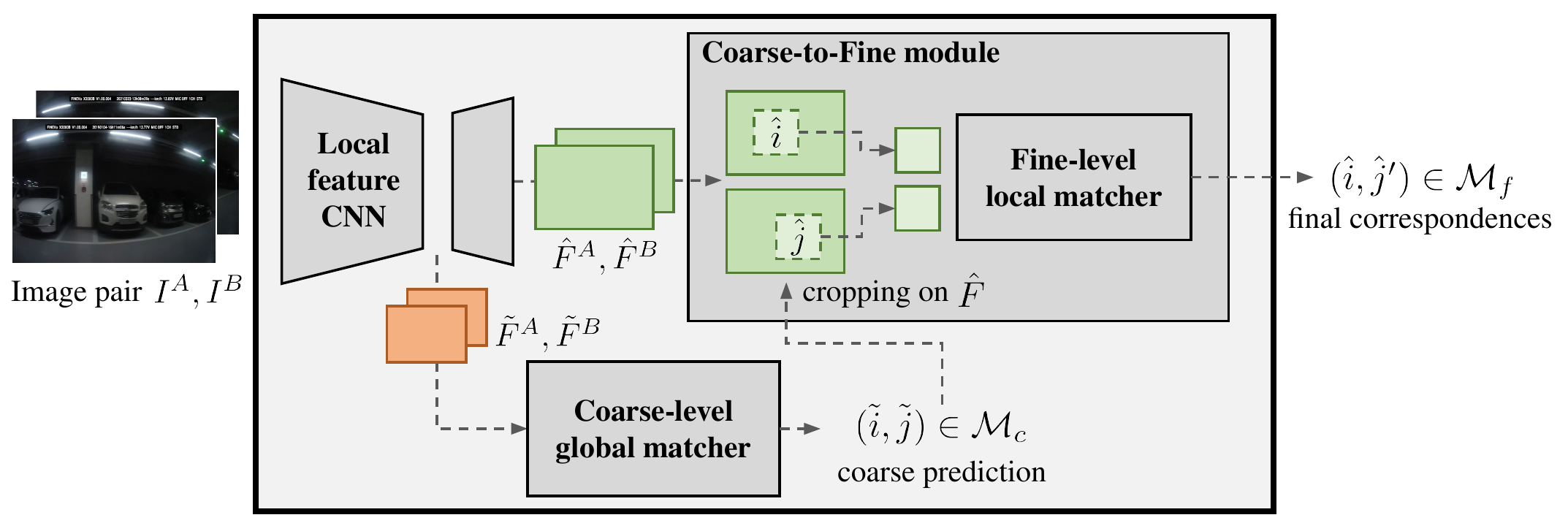}
    \caption{
    \textbf{Overview of the deep local feature matching architecture.} The local feature CNN produces feature maps from an image pair. The coarse-level global matcher produces coarse correspondences between all coarse grid points. The coarse-to-fine module narrows down the search space of the fine-level local matcher.
    }
    \label{fig:matching}
\end{center}
\end{figure*}

\fref{fig:matching} illustrates the procedure of deep dense local feature matching. First, LoFTR computes coarse-level feature maps $\tilde{F^A}$ and $\tilde{F^B})$, and fine-level feature maps $\hat{F^A}$ and $\hat{F^B}$, given a pair of images $I^A$ and $I^B$, respectively, using a CNN. A coarse-level global matcher predicts coarse correspondences $\mathcal{M_c}$ from $\tilde{F^A}$ and $\tilde{F^B}$. For every coarse correspondence $(\tilde{i}, \tilde{j}) \in \mathcal{M_c}$, its local patches are cropped from the fine-level feature maps. Then, the final matcher produces fine-grained matches $(\hat{i}, \hat{j}) \in \mathcal{M_f}$ by refining their coordinates to subpixel level.

The coarse-level global matcher computes the similarity between all pairs of locations and produces categorical probability distributions by applying softmax functions. A pair is considered as a correspondence if the points are mutual nearest neighbors and the probability is greater than a threshold. 

The final matcher computes the correlation between a central vector of a cropped window and all vectors of another cropped window to produce a heatmap that represents the matching probability of each location. Then the sub-pixel coordinates are obtained by expectation over the heatmap.

The above procedure is efficient and robust since it resembles the human perception that reflects surroundings for finding correspondences. Hence, we count the number of correspondences between a query and pre-collected images to consider the image with the most correspondences as the match.

\begin{figure*}[t]
\begin{center}
\includegraphics[width=0.65\linewidth]{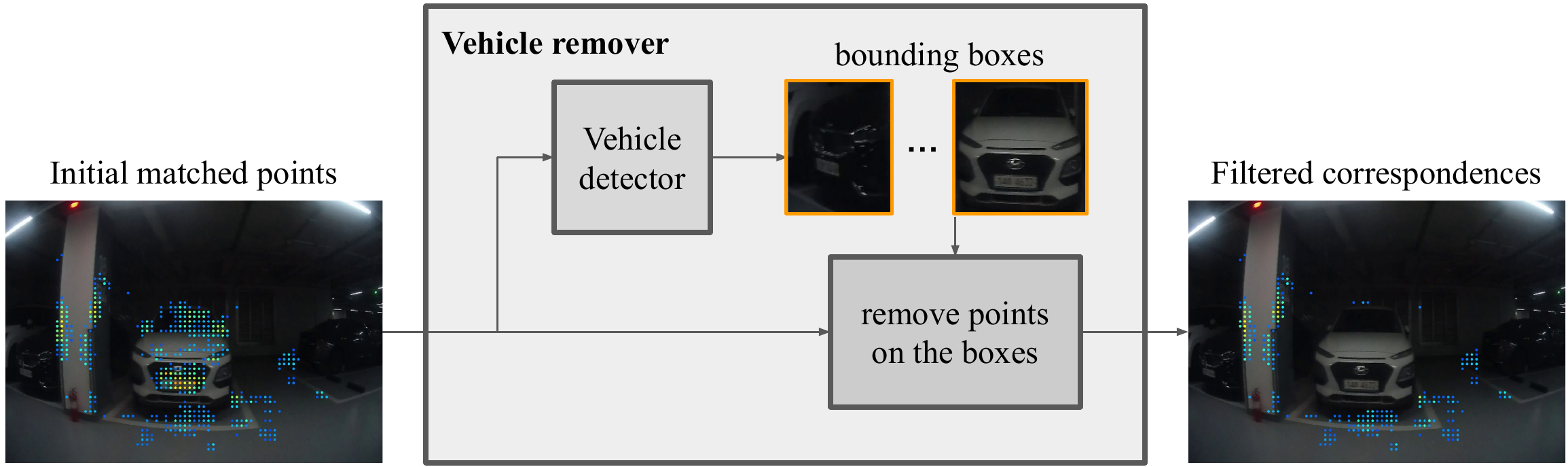}
\caption{
\textbf{Overview of the vehicle removal process.} The off-the-shelf vehicle detector produces bounding boxes of vehicles. Then the correspondences on the boxes are removed.
}
\label{fig:removal}
\end{center}
\end{figure*}

\subsection{Vehicle removal}
Even if we assume that the deep dense local feature matching finds all visual correspondences between images correctly, the correspondences from vehicles lead to false matches because there can be identical or similar-looking vehicles between false matches. Furthermore, vehicles are not fixed to specific locations; matches on the vehicles may not contribute to the localization.

Hence, we remove the correspondences found on vehicles using an off-the-shelf object detector \cite{girshick2015fast, lin2017focal, redmon2016you, yolov5} as shown in \fref{fig:removal}. The detector finds bounding boxes of vehicles so that the points in the boxes are ignored. This post-processing improves the performance and robustness of our framework.

\paragraph{Choice of the vehicle detector}

We choose YOLOv5 \cite{yolov5} as our vehicle detector. The smaller models (YOLOv5s and YOLOv5m) are disregarded due to their inaccuracies, \eg, detecting one vehicle among three vehicles in a scene. Larger models (YOLOv5x6 and YOLOv5l6) are disregarded due to their misclassifications.

\subsection{Localization criterion}

Given the filtered correspondences between the query image and all pre-collected images, we count the number of correspondences and output the section with the highest number of correspondences.

\section{Experiments}
In this section, we demonstrate the effectiveness of our method on the new dataset collected from a parking lot.

\subsection{Dataset}
\begin{figure}[t]
\includegraphics[width=\linewidth]{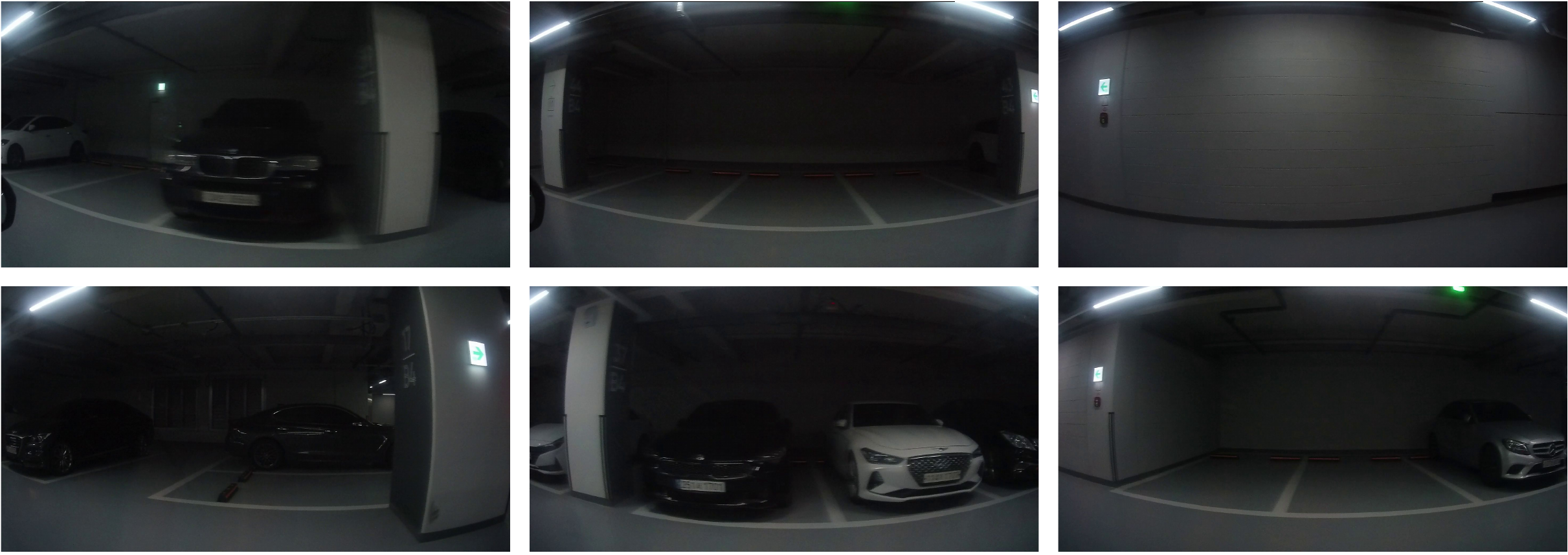}
\caption{
\textbf{Subset of images in the collected dataset.} The scenes contain textureless walls and similar-looking vehicles.
}
\label{fig:dataset}
\end{figure}

In order to evaluate our framework, we collect a set of images from a vehicle with side-looking cameras, slowly driving through the aisles. We run the collecting drive twice: one for the pre-collected images and one for the queries. Although the ideal procedure is to take a front-looking image from each parked location, it immensely increases the time to collect the images. Collecting images on a driving car makes the practice much easier. Example images are shown in \fref{fig:dataset}. Each query may have up to two ground truth labels because there are ambiguous cases where the query image covers two sections (last row in \fref{fig:dataset}).

%
\begin{table}[!t]
\renewcommand{\arraystretch}{1.3}
\caption{Statistics of the collected dataset}
\label{tab:dataset}
\centering
\begin{tabular}{|c|c|}
\hline
                     & count \\ \hline
pre-collected images & 267   \\ \hline
query images         & 99    \\ \hline
sections             & 36    \\ \hline
\end{tabular}
\end{table}

\begin{figure*}[t]
    \includegraphics[width=\linewidth]{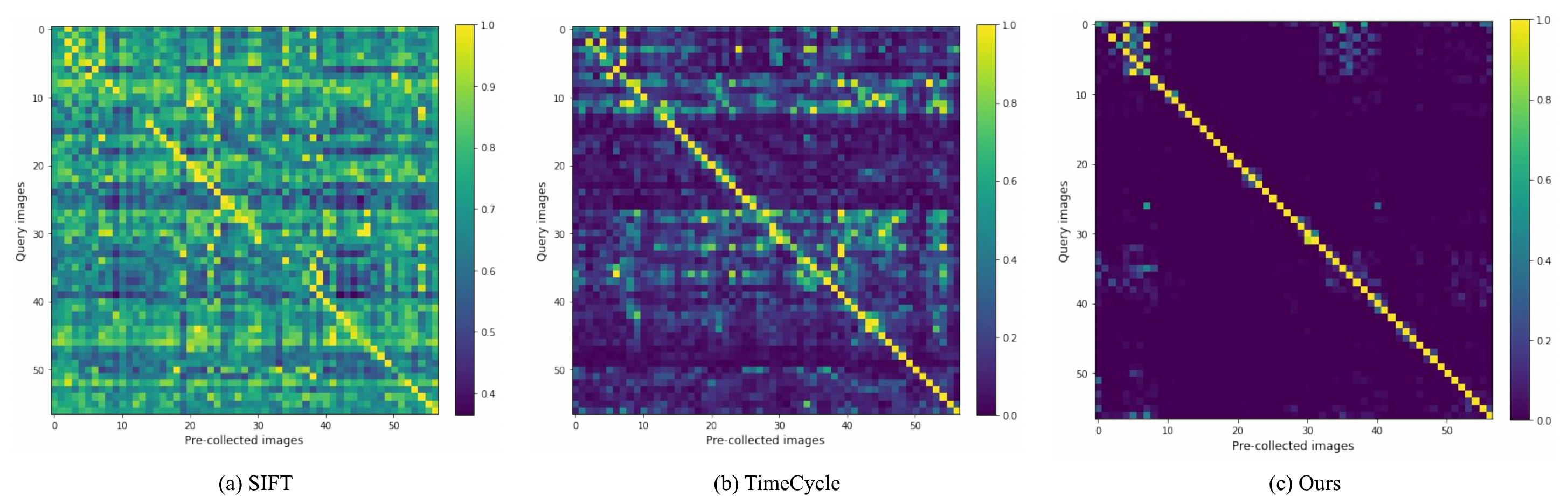}
    \caption{
    \textbf{Visualization of the number of correspondences.} An element in the matrices represents the number of local correspondences between a pair of the query image and pre-collected image. The diagonal elements show the number of local correspondences between ground truth image pairs. We divide each row by the maximum for normalization. 
    }
    \label{fig:confmat}
\end{figure*}

\begin{figure}[t]
\centering
    \includegraphics[scale=0.5]{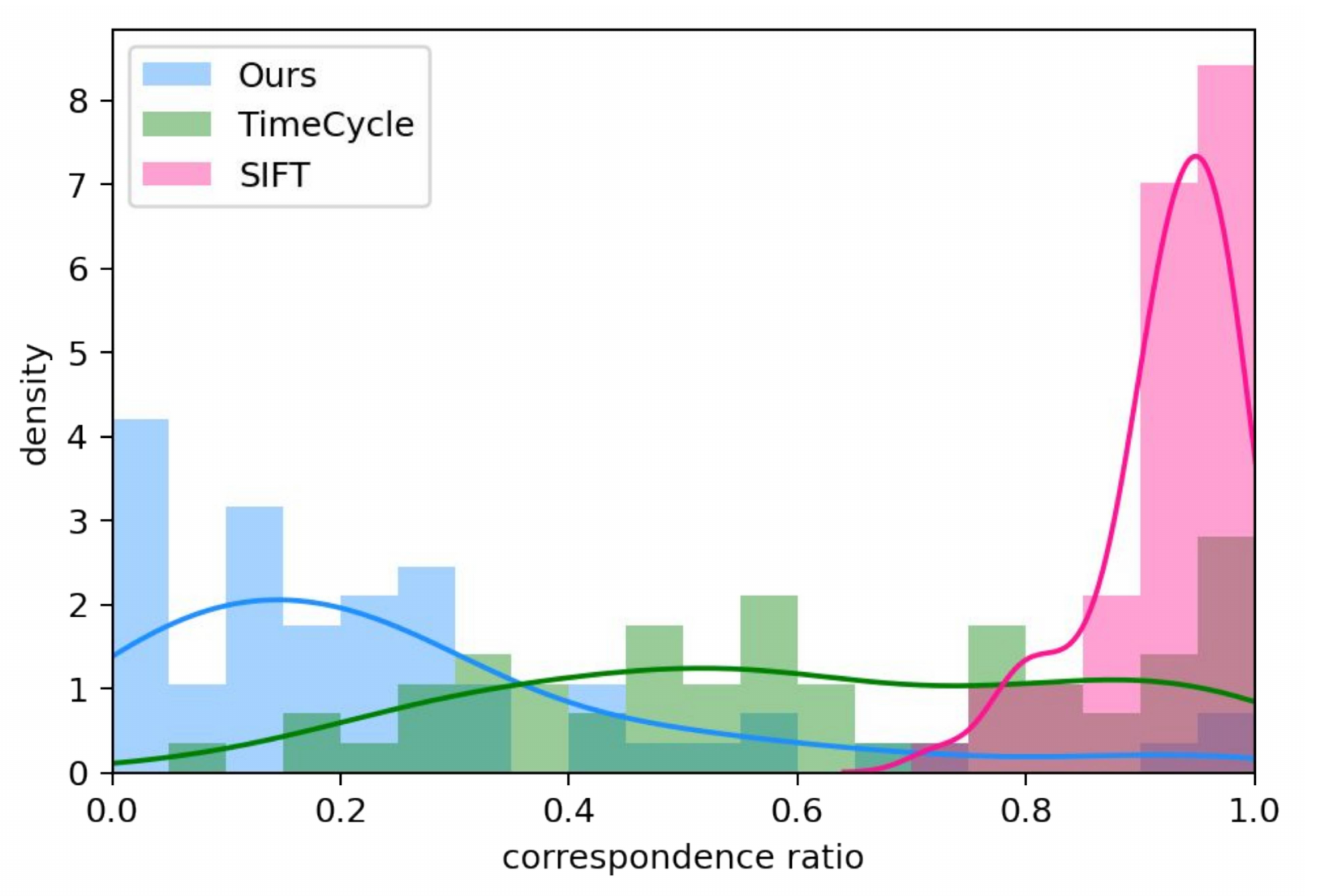}
    \caption{\textbf{Vulnerability of the methods due to false correspondences.} We compute the ratio of the number of correspondences of the second-best pair to the number of correspondences of the best pair. Then the higher ratio implies that the method tends to confuse the best and the second-best images. The bars denote the number of samples in their bins and the curve denotes the probability density function. The lower the ratio, the better the distinction between the first and second closest regions.}
    \label{fig:hist}
\end{figure}
\subsection{Competitors}

We consider three alternatives for matching-based image retrieval: a traditional handcrafted feature (SIFT) \cite{lowe2004distinctive}, correspondence learned from videos (TimeCycle) \cite{wang2019learning}, and a place recognition approach (NetVLAD) \cite{arandjelovic2016netvlad}. All methods are tested using the public implementations \footnote{\url{https://opencv24-python-tutorials.readthedocs.io/en/stable/py_tutorials/py_feature2d/py_sift_intro/py_sift_intro.html}}\footnote{\url{https://github.com/xiaolonw/TimeCycle}}\footnote{\url{https://github.com/Nanne/pytorch-NetVlad}}.

\subsection{Evaluation metrics}

We evaluate the performances as the accuracy of predicting correct locations. A query can be annotated up to two sections reflecting possible ambiguous cases where a query is taken in the middle of two sections. Thus, a prediction given a query is considered correct if its prediction is in the ground truth labels.

\begin{table}[t]
\renewcommand{\arraystretch}{1.3}
\caption{Accuracy comparison over different matching methods}
\label{tab:quant}
\centering
\begin{tabular}{|c|c|}
\hline
          & accuracy \\ \hline
SIFT      & 9.1\%    \\ \hline
TimeCycle & 66.7\%    \\ \hline
NetVLAD   & 83.8\%    \\ \hline
Ours      & 86.9\%    \\ \hline
\end{tabular}
\end{table}
\begin{figure*}[t]
\includegraphics[width=\linewidth]{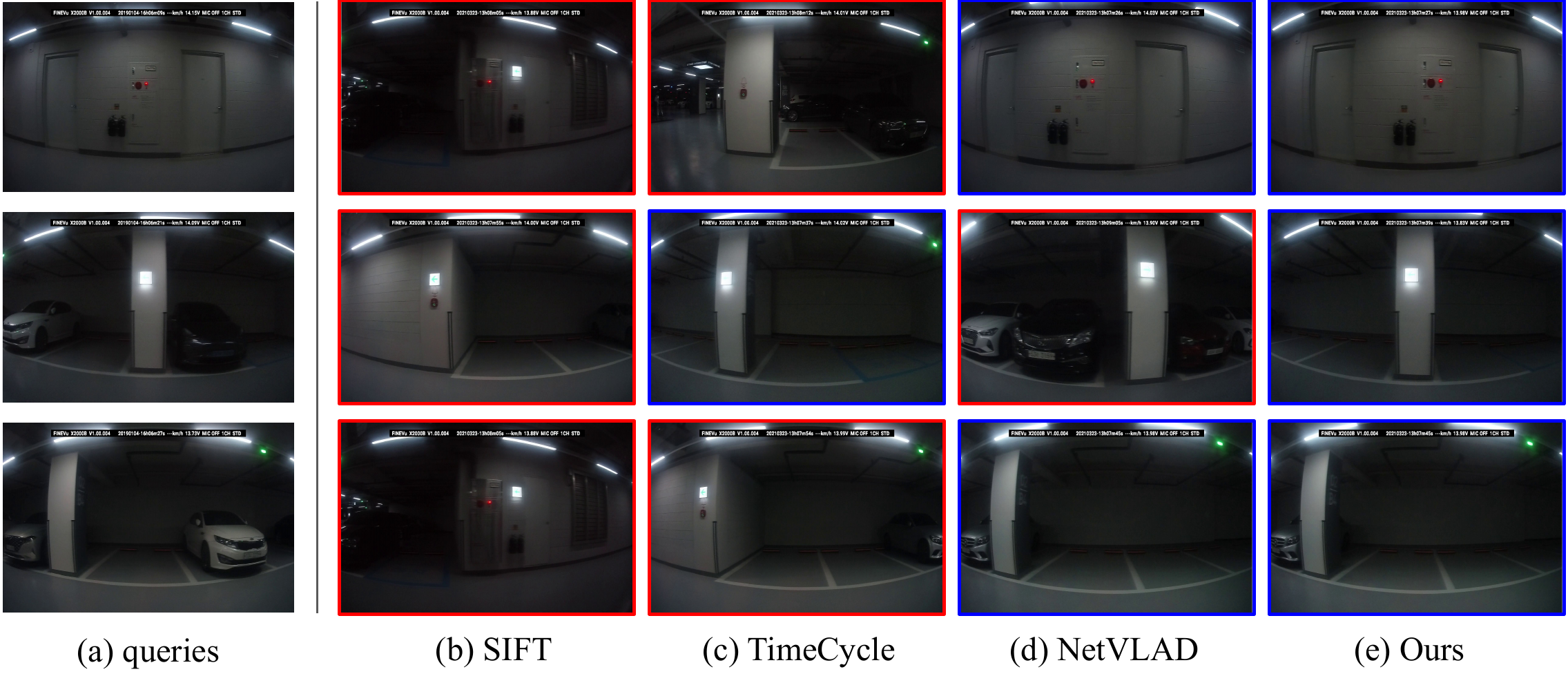}
\caption{
\textbf{Qualitative comparison across the competitors.} Our method produces correct matches while other methods struggle with different challenges.
}
\label{fig:qual}
\end{figure*}
\begin{figure*}[t]
\includegraphics[width=\linewidth]{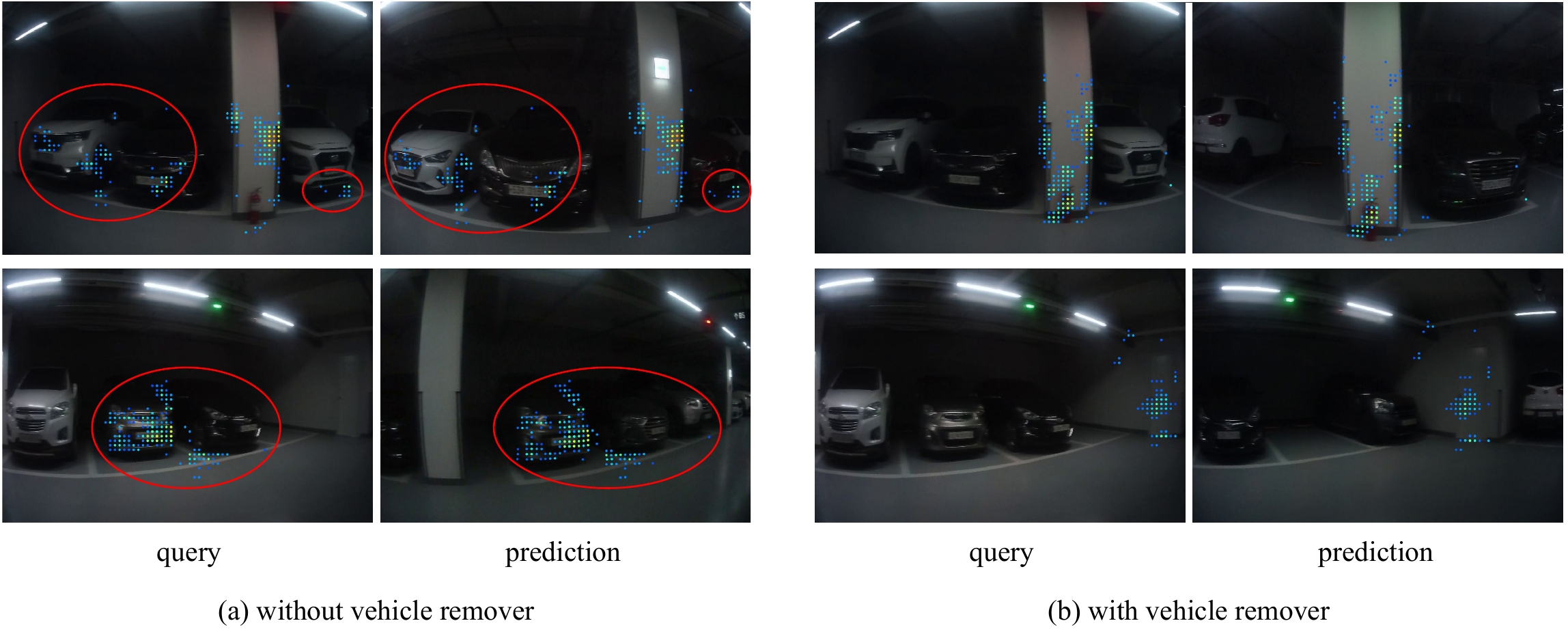}
\caption{
\textbf{Examples of vehicle removal.} (a) Identical or similar-looking vehicles may produce false matches. (b) Removing the correspondences on the vehicles leads the correct images to win the most correspondences.
}
\label{fig:remover}
\end{figure*}

\subsection{Quantitative comparison}

\tref{tab:quant} shows the accuracies of the competitors. The interest-point-based method (SIFT) struggles as expected. We suppose that 1) the interest points are hardly detectable on the textureless walls and pillars, and 2) the correspondences on the identical cars in different locations misleads the localization. Although TimeCycle is trained with reasonable supervision for dense correspondence, still, textureless walls make the cycle-consistency ambiguous so that the correspondences are unstable. NetVLAD achieves the closest performance to ours. We attribute the loss of accuracy to the similar-looking objects on which NetVLAD relies. Lastly, our choice achieves the best performance among the competitors. It is due to the coarse-to-fine nature of LoFTR \cite{sun2021loftr}.

\subsection{Analysis}

\fref{fig:confmat} illustrates the number of matches between the query images and pre-collected images. We manually collected 57 ground truth image pairs from two different sets of collected images, \ie, the image pairs capture the same place with similar viewpoints. Since the maximum number of matches varies considerably depending on the query image, we divide each row of the matrix by its maximum score for easy comparison. Furthermore, \fref{fig:hist} illustrates histograms of the ratio of the second-most matches to the most matches of each query image. Our approach, marked in blue, tends to exhibit low ratios, meaning that it is more effective than competitors in distinguishing similar-looking different places. Hence, relying on the number of matches between pairs is a viable approach.

\subsection{Qualitative comparison}

\fref{fig:qual} illustrates differences across the competitors. SIFT completely fail to find correspondences. TimeCycle finds only small number of correspondences on poorly textured walls and floors. NetVLAD achieves reasonable performance but is fooled by a similar-looking exit indicator on the pillar. Our choice shows robust performance in most cases thanks to the robustness of LoFTR.

\subsection{Effectiveness of the vehicle remover}

\tref{tab:ablation} supports the necessity of the vehicle remover by comparing accuracies of our framework with and without the vehicle remover. Without the vehicle remover, the accuracy falls by 2.1\%p. \fref{fig:remover} illustrates examples where vehicle remover is effective. The gap will be larger if there are more vehicles parked in the parking lot. 

\begin{table}[t]
\renewcommand{\arraystretch}{1.3}
\caption{Necessity of the vehicle remover.}
\label{tab:ablation}
\centering
\begin{tabular}{|c|c|}
\hline
                             & accuracy \\ \hline
Ours without vehicle remover & 0.848    \\ \hline
Ours full                    & 0.869    \\ \hline
\end{tabular}
\end{table}
\begin{figure*}[t]
    \includegraphics[width=\linewidth]{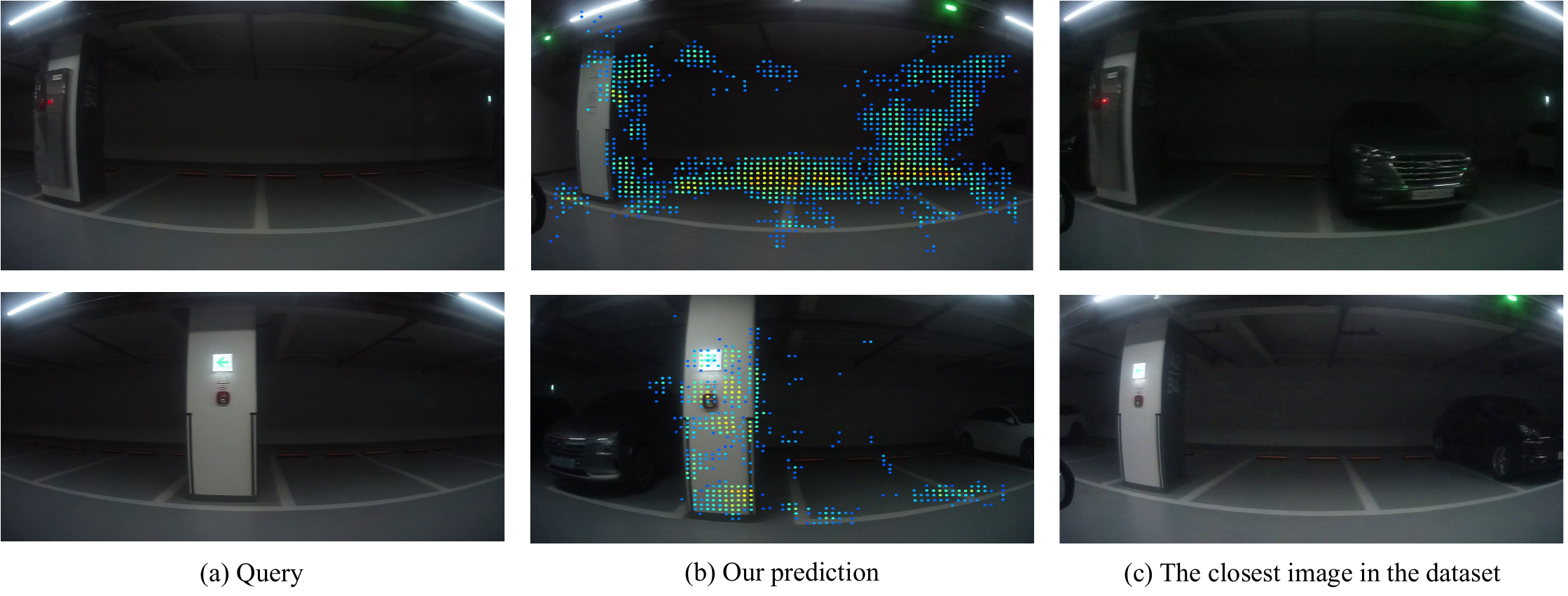}
    \caption{
    \textbf{Failure cases.} Our framework makes mistakes when an overwhelming number of correspondences occur. It is a common pitfall of correspondence-based approaches in environments where similar scenes repeatedly exist. Rejecting the images with differences may improve the performance.
    }
    \label{fig:failure}
\end{figure*}

\subsection{Failure cases}
As our method relies on local correspondences, it may fail when false correspondences occur. For example, there are many similar-looking regions on dark walls or floors (\fref{fig:failure}). When the number of such false correspondences overwhelms the number of correct matches, our method fails because it assigns the query image to the image with the most number of correspondences. While it is a common pitfall of correspondence-based approaches, rejecting the correspondences with different geometry or rejecting the images with different appearances may improve the performance.

\section{Conclusion}
In this paper, we proposed a framework for visual localization, especially for indoor parking lot environments. Indoor parking lots are challenging environments because the walls are poorly textured and identical vehicles are confusing. Our framework consists of carefully chosen components: deep dense local feature matcher and vehicle remover with an off-the-shelf detector. LoFTR robustly finds correspondences while alternatives struggle on poorly textured walls or similar-looking objects. The vehicle remover effectively disregards spurious matches on vehicles to alleviate confusion due to identical vehicles. Experiments on a real-world dataset quantitatively and qualitatively validate the superiority of our framework upon alternatives. Our framework can be further improved by employing a sequence of query images to the final location or physical sensors such as an accelerometer. 


%



\section*{Acknowledgment}
This work is collaborated by Jeongmin Bae, Sanghuk Lee, and Youngjung Uh. Jeongmin Bae, Sanghuk Lee, and Youngjung Uh are with the Department of Artificial Intelligence at Yonsei University. We highlight that this manuscript aims to share an empirical cutting-edge use case of visual localization in indoor environments rather than suggesting academic contribution.

\ifCLASSOPTIONcaptionsoff
  \newpage
\fi



\newpage
\bibliographystyle{IEEEtran}
\bibliography{references}
\end{document}